%% file: main.tex
\pdfoutput=1

\documentclass[11pt]{article}

\usepackage[]{emnlp2021}

\usepackage{times}
\usepackage{latexsym}

\usepackage{multirow}
\usepackage{multicol}

\usepackage[utf8]{inputenc}
\usepackage{bbold}
\usepackage{url,graphicx,tabularx,array, booktabs, mathrsfs, color, epstopdf, tabu, svg, diagbox, mathtools, multirow} 
\usepackage{amsmath, amssymb, bm, cases, enumitem, bm}
\usepackage{dsfont}

\usepackage[T1]{fontenc}

\usepackage[utf8]{inputenc}

\usepackage{microtype}

\usepackage{graphicx,amssymb,algorithm,algpseudocode,amsmath}

\title{Self-training Improves Pre-training for Few-shot Learning in Task-oriented Dialog Systems}

\author{
Fei Mi$^1$,
Wanhao Zhou$^2$,
Fengyu Cai$^2$,
Lingjing Kong$^2$,
Minlie Huang$^{3}$ \and
Boi Faltings$^2$\\
$^1$Huawei Noah's Ark Lab\\
$^2$LIA, EPFL\\
$^3$CoAI, DCST, Tsinghua University
\\
\texttt{mifei2@huawei.com}, \quad
\texttt{aihuang@tsinghua.edu.cn} \\
\texttt{\{wanhao.zhou,fengyu.cai,lingjing.kong,boi.faltings\}@epfl.ch}
} 

\begin{document}
\maketitle
\begin{abstract}
As the labeling cost for different modules in task-oriented dialog (ToD) systems is expensive, a major challenge is to train different modules with the least amount of labeled data. Recently, large-scale pre-trained language models, have shown promising results for few-shot learning in ToD. In this paper, we devise a self-training approach to utilize the abundant unlabeled dialog data to further improve state-of-the-art pre-trained models in few-shot learning scenarios for ToD systems.
Specifically, we propose a self-training approach that iteratively labels the most confident unlabeled data to train a stronger \emph{Student} model. Moreover, a new text augmentation technique (GradAug) is proposed to better train the \emph{Student} by replacing non-crucial tokens using a masked language model.
We conduct extensive experiments and present analyses on four downstream tasks in ToD, including intent classification, dialog state tracking, dialog act prediction, and response selection. Empirical results demonstrate that the proposed self-training approach \emph{consistently} improves state-of-the-art pre-trained models (BERT, ToD-BERT) when only a small number of labeled data are available.
\end{abstract}

\input{text/intro}

\input{text/related}

\input{text/model}

\input{text/exp}

\input{text/conclusion}

\newpage
\bibliographystyle{acl_natbib}
\bibliography{naacl}
\input{tables/dataset-setting}

\newpage
\input{text/appendix}



\end{document}

%% file: text/intro.tex
\section{Introduction}

Large-scale pre-trained language models, such as BERT \cite{devlin2018bert}, UniLM \cite{DBLP:conf/nips/00040WWLWGZH19}, GPT \cite{radford2018improving}, GPT-2 \cite{radford2019language}, and GPT-3 \cite{brown2020language}, have shown great few-shot or zero-shot learning abilities in various NLP tasks with the help of task-agnostic language knowledge learned via pre-training tasks.
In task-oriented dialog (ToD) systems, the labeling cost is very high such that the size of well-labeled data is often small. Therefore, few-shot learning in ToD is important and valuable in many practical applications. Many attempts \cite{peng2020few,peng2020soloist,wu2020tod} have been proposed to leverage large-scale pre-trained language models to improve few-shot learning in ToD. Specifically, a model pre-trained on general text corpora is further trained on public ToD datasets. 

Although the size of labeled data is often small, a practical ToD system de facto has many unlabeled dialog data. Therefore, utilizing unlabeled data to improve a ToD system is practically important.
In this paper, we take a semi-supervised self-training (ST) perspective to iteratively train a better \emph{Student} model using unlabeled data \cite{scudder1965probability,yarowsky1995unsupervised}. 
ST has been successfully applied to a variety of tasks, including image classification \cite{yalniz2019billion,xie2020self,zoph2020rethinking}, automatic speech classification \cite{synnaeve2019end,kahn2020self,park2020improved,likhomanenko2020slimipl}, sequence generation \cite{he2019revisiting}, and natural language understanding \cite{du2020self}.

We are going to study this research question: \emph{can self-training provide complementary benefits on top of the strong pre-training models for few-shot learning in ToD?} Recently,
\citet{xie2020self,zoph2020rethinking} studied a similar question in the
context of image classification, showing that ST effectively refines pre-training models. 
\citet{du2020self} also recently showed the benefit of ST over pre-training for general natural language understanding. Yet, their main proposal is to crawl a large amount of similar unlabeled data from the web.

In this paper, we propose a self-training approach based on iterative pseudo-labeling \cite{lee2013pseudo}. It first trains a \emph{Teacher} on the labeled samples. The \emph{Teacher} then iteratively generates pseudo-labels for the most confident subset of unlabeled samples to train a better \emph{Student}. 
To train a more robust \emph{Student} during self-training, we propose a data augmentation technique called GradAug. GradAug first ``masks'' a fraction of tokens of a dialog input. Then, it reconstructs the corrupted text with a pre-trained masked language model of BERT.
Different from \citet{ng2020ssmba}, the probability of masking a token is conditioned on the gradient of the corresponding token embedding w.r.t. the downstream task. In this way, GradAug prevents replacing tokens that are critical for a downstream task. 

The main contribution of this paper is three-fold:
\begin{itemize}[itemsep=0pt,topsep=2pt,leftmargin=12pt]
\item This is the first attempt to study the effect of self-training on top of existing strong pre-trained models for ToD in few-shot learning scenarios.

\item We propose a self-training method to gradually train a stronger \emph{Student} by iteratively labeling the most confident unlabeled data and a new text augmentation technique (GradAug).

\item We conduct extensive experiments on four downstream tasks in ToD,  including intent classification, dialog state tracking, dialog act prediction, and response selection. Empirical results demonstrate that self-training \textit{consistently} improves state-of-the-art pre-trained models (BERT, ToD-BERT \citet{wu2020tod}).

\end{itemize}

%% file: text/related.tex
\section{Related Work}

\subsection{Pre-training for ToD Systems}

\citet{budzianowski2019hello} first applied GPT-2 to
train a response generation model by taking the system belief state, database entries, and last dialog turn as input.
\citet{henderson2019training} pre-trained a response selection model for ToD by first pre-training on general-domain conversational corpora (Reddit).
\citet{ham2020end} trained the pre-trained GPT-2 for dialog state tracking and response generation on MultiWOZ \cite{budzianowski2018multiwoz}.
\citet{hosseini2020simple} proposed SimpleToD to train the pre-trained GPT-2 on three different sub-tasks (dialog state tracking, dialog act prediction, and response generation) of ToD as a sequence prediction problem. 

Recent studies have shown that large-scale pre-trained language models are good few-shot learners \cite{brown2020language}. Several studies have also confirmed these findings for ToD. For the task of generating responses conditioned on a semantic representation, GPT-2 was leveraged by \citet{peng2020few} to improve few-shot learning and by \citet{mi2020continual} in a continual learning setting.
\citet{peng2020soloist} utilized GPT-2 for end-to-end response generation from dialog contexts in a few-shot learning scenario. 
\citet{wu2020tod} further trained a BERT model on multiple ToD corpora to improve few-shot learning performance on four different downstream tasks. 

\subsection{Self-training}
\label{sec:related_st}

The first focus of self-training is designing better policies to label unlabeled samples. \citet{zhang2011cotrade} evaluated the
confidence via a statistic-based data editing technique. \citet{lee2013pseudo} designed an annealing function that gradually increases the loss of labeled samples during training. \citet{amiri2019neural} utilized a Leitner queue \cite{dempster1989spacing} to gradually put confident samples in the front. \citet{niu2020self} selected the most confident samples with prediction loss below some threshold. 
\citet{kumar2010self,ma2017self,li2019learning,mukherjee2020uncertainty} proposed to learn sampling weights for unlabeled data to control the selection process. 
Reinforcement learning (RL) methods \cite {chen2018learning,wu2018reinforced,ye2020zero} designed an additional Q-agent as the sample selector.
Nevertheless, methods using learnable weights or RL provide marginal benefits compared to the elevated optimization cost. 
As designing new sample selection schemes is not our primary focus, we will go for a simple and effective pipeline described in Section \ref{subsec:ST_algo}.  Specialized explorations on this topic are orthogonal to the focus of this paper and will be left as future work.

The second focus of self-training is to improve the robustness of the \emph{Student} model trained from potentially noisy pseudo-labeled samples.
Data augmentation techniques are widely used. In computer vision, recent works demonstrated the benefit of different stochastic augmentation tricks, including input transformations \cite{laine2016temporal,xie2020self,zoph2020rethinking}, dropout \cite{laine2016temporal,xie2020self,zoph2020rethinking}, adversarial samples \cite{miyato2018virtual}, and Mixup \cite{berthelot2019mixmatch,berthelot2019remixmatch}.
Text augmentation is more challenging because of the complex syntactic and semantic structures.
\citet{miyato2016adversarial} utilized adversarial training to apply perturbations to word embeddings.
\citet{wei2019eda} proposed EDA using basic synonym replacement, random insertion, swap, and deletion.
\citet{kumar2019submodular} proposed to maximize a monotone sub-modular function to obtain diverse paraphrases. 
\citet{xie2019unsupervised} proposed UDA applying back-translation \cite{edunov2018understanding} and word replacement using a Tf-Idf metric.
\citet{he2019revisiting} studied the effect of dropout compared to back-translation during self-training for the neural sequence generation task.
\citet{chen2020mixtext} proposed MixText that utilizes Manifold Mixup \cite{verma2019manifold} to interpolate hidden layers corresponding to semantic representations of BERT.
\citet{ng2020ssmba} proposed SSMBA utilizing the masked language model of BERT to replace words.
In experiments, we compare the proposed GradAug technique with state-of-the-art text augmentation methods.

%% file: text/model.tex
\section{Background of Using Pre-trained Models for Downstream Tasks in ToD}

In this section, we first briefly overview the pipeline of utilizing large-scale pre-trained models for four common downstream tasks (intent classification, dialog state tracking, dialog act prediction, and response selection) in ToD. We denote the input and label of different downstream tasks as $x$ and $y$, and a prediction model is denoted as $\hat{y}_x = F(x)$. $F$ can often be decomposed into two parts. The first part is a \emph{feature extractor} $\bm{h}=A(x) \in \mathbb{R}^l$ 
which computes a hidden representation $\bm h$ of $x$, and the second part is an \emph{output network} for prediction. 
Large-scale pre-trained language models serve as \emph{feature extractor} $A$ to compute a hidden representation for an input. For example, we use the [CLS] embedding of BERT as the hidden representation $\bm{h}$ when BERT is adopted as $A$. Different \emph{output networks} are designed for different downstream tasks, and the details following ToD-BERT \cite{wu2020tod} are described below. 

\paragraph{Intent classification.} This is a multi-class classification problem to predict the single intent label $y$ of an input utterance $x$. The model computes the probability over $I$ possible intents as:
\begin{equation}
\bm{p}_{int} = Softmax(W_1 \cdot  A(x)) \in \mathbb{R}^I,
\end{equation}
where $W_1 \in \mathbb{R}^{I \times l}$ is a trainable weight matrix, and the model is optimized by the standard cross-entropy loss compared to the ground truth. 

\paragraph{Dialog state tracking.} It is a multi-class classification problem based on a predefined ontology. Unlike intent classification, the dialog history (a sequence of utterances) is used as the input $x$. For each (domain, slot) pair, the model predicts a score over all potential slot values. For the $i$-th slot value $v_i^j$ of the $j$-th pair, the cosine similarity score compared to the input $x$ is computed as follows:
\begin{equation}
s_i^j = Cosine(G_j(A(x)), A(v_i^j)) \in \mathbb{R}^1,
\end{equation}
where $G_j$ is the slot projection layer of the $j$-th pair, and the number of layers $|G|$ equals the number of (domain, slot) pairs. The model is trained with the cross-entropy loss summed over all the pairs.

\paragraph{Dialog act prediction.} 
This is a multi-label classification problem to predict dialog act (DA) intents for the next system response. The model takes a dialog history as input $x$ and predicts a Bernoulli outcome for each possible DA intent as:
\begin{equation}
\bm{a} = Sigmoid(W_2 \cdot A(x)) \in \mathbb{R}^N,
\end{equation}
where $W_2 \in \mathbb{R}^{N \times l}$ is a trainable weight matrix, and $N$ is the number of possible DA intents. Values in $\bm{a}$ are between $[0,1]$, and the model is optimized by a binary cross-entropy loss w.r.t. the ground truth. A threshold of 0.5 is applied during inference.

\paragraph{Response selection.} This task predicts the most relevant system response from a candidate pool. A dual-encoder model \cite{henderson2019training} is adopted to compute the
 similarity between the input dialog history $x$ and the $i$-th candidate response $c_i$:
\begin{equation}
r_i = Cosine(A(x), A(c_i)) \in \mathbb{R}^1.
\end{equation}
During training, we randomly sample 20 negative responses for each ground truth response. 
A cross-entropy loss is applied aiming to rank the ground truth highest.

\section{Self-training}
\label{sec:self_training}

 In this section, we introduce our self-training (ST) algorithm. 
 The overall ST algorithm is introduced in Section \ref{subsec:ST_algo}, and a new text augmentation method (GradAug) for ST to train a more robust \emph{Student} is elaborated in Section \ref{subsec:GradAug}.
 
 \subsection{Overall ST Algorithm}
\label{subsec:ST_algo}
 During training, two data pools are maintained and denoted as $U$
(unlabeled data) and $L$ (labeled data)
. Two versions of the model are maintained, \emph{Teacher} ($F^T$) and \emph{Student} ($F^S$). Before the iterations of ST start, the \emph{Teacher} is first trained on the initial small number of labeled data $L$ to ``warm up''.

 \begin{figure}[t!]
		\centering
		\includegraphics[width=0.495\textwidth]{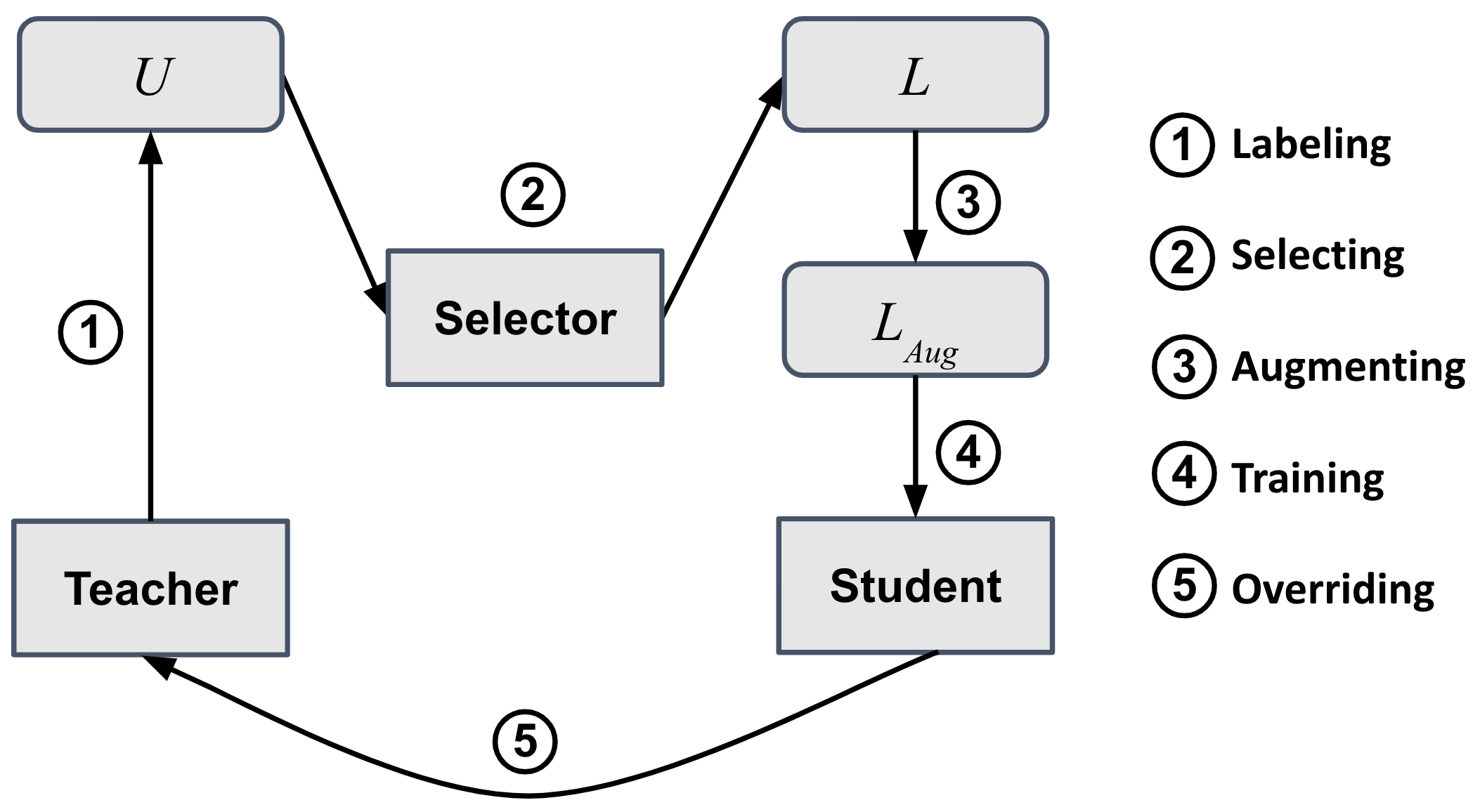}
		\caption{Pipeline of one ST iteration. The \emph{Teacher} first generates predictions for data in $U$. Then, the \emph{Selector} chooses the most confident samples based on the \emph{Teacher}'s predictions and assign pseudo labels to them before appending to $L$. Afterwards, $L$ is augmented by ``GradAug'' to train a \emph{Student}. Lastly, the trained \emph{Student} becomes the \emph{Teacher} in the next iteration. Multiple iterations are computed till the \emph{Student} converges.} \vspace{-0.05in}
		\label{fig:pl_pipeline}
	\end{figure}

\paragraph{Pseudo-Labeling.} In the beginning of an ST iteration, the \emph{Teacher} first makes predictions on $U$.
For every data input $x \in U$, the \emph{Teacher} predicts the label of $x$ as $\hat{y}_x = F^T(x)$. We set the predicted score of the prediction $\hat{y}_x$ as the \emph{confidence score} $s_x$ for this prediction. When there is only a single label in the prediction $\hat{y}_x$ (c.f. intent classification, response selection), $s_x$ is the prediction score corresponding to the predicted label. When there are multiple labels in the prediction $\hat{y}_x$ (c.f. dialog state tracking, dialog act prediction), $s_x$ takes the mean of the prediction scores corresponding to the predicted labels. In each iteration, the \emph{Selector} chooses top-$k$ 
instances from $U$ with the highest confidence scores, and assigns the corresponding predictions $\hat{y}_x$ as labels to them. These labeled instances will be moved from $U$ to $L$. 

\paragraph{Iterative \emph{Student} training.} The updated $L$ is used to train a stronger \emph{Student} model. We applied dropout \cite{srivastava2014dropout} and a new text augmentation technique (GradAug) introduced later in Section \ref{subsec:GradAug} which augments $L$ to $L_{Aug}$.
At the end of each iteration, the \emph{Teacher} model is overridden by the current \emph{Student} to be used in the next iteration.
We reinitialize the \emph{Student} in every iteration to avoid over-fitting the initial and earlier data in $L$ in multiple training iterations.
As noted by \citet{xie2020self,du2020self}, the \emph{Student} should have an equal or larger capacity than the \emph{Teacher} to gradually learn from $L$ with increasing size. In this paper, we set the \emph{Student} the same size as the \emph{Teacher}, and we demonstrate in experiments that consistent improvements can be achieved without increasing model capacity.

Details of our ST algorithm are described in Algorithm \ref{algo:ST}, and the pipeline of one ST iteration (i.e., the ``While'' loop in Algorithm \ref{algo:ST}) is visualized in Figure \ref{fig:pl_pipeline}. 

\renewcommand\algorithmicrequire{\textbf{Input:}}
\renewcommand\algorithmicensure{\textbf{Output:}}
\begin{algorithm}[!t]
\caption{Self-training (ST) for ToD}\label{algo:ST}
\begin{algorithmic}[1]
\Require{Labeled data: $L$, Unlabeled data: $U$, Teacher: $F^T$, Student: $F^S$, Number of pseudo-labeled data in an iteration: $k$, Number of augmentations per input: $q$ }
\Ensure{A trained Student $F^S$}
\State Initialize $F^T$ and train $F^T$ on $L$
\While{$F^S$ not good enough $\And U \neq$ \O}
\State Initialize $F^S$, $L' \gets$ $Priority\_list()$ 
\For{$x \in U$}
\State Compute prediction label $\hat{y}_x = F^T(x)$ 
\State Compute confidence score $s_x$
\State $L'.insert(\{x, \hat{y}_x, s_x\})$
\EndFor
\State $L' \gets L'.top(k)$ 
\State $L \gets L \cup L'$, $U \gets U \backslash L'$
\State $L_{Aug} \gets GradAug(L, F^T, q)$
\State Train $F^S$ on $L_{Aug}$ with dropout
\State $F^T \gets F^S$
\EndWhile
\end{algorithmic}
\end{algorithm}

\subsection{Text Augmentation (GradAug)}
\label{subsec:GradAug}
Next, we propose a novel text augmentation technique called ``GradAug'' for data in $L$ to train a more robust \emph{Student}. Our method employs the masked language model (MLM, \citet{devlin2018bert,liu2019roberta}), which is a common pre-training strategy for BERT-like architectures. In MLM, some tokens are replaced by the special token [MASK], and the model is asked to reconstruct the original tokens from the context. 

\begin{figure}[t!]
		\centering
		\includegraphics[width=0.46\textwidth]{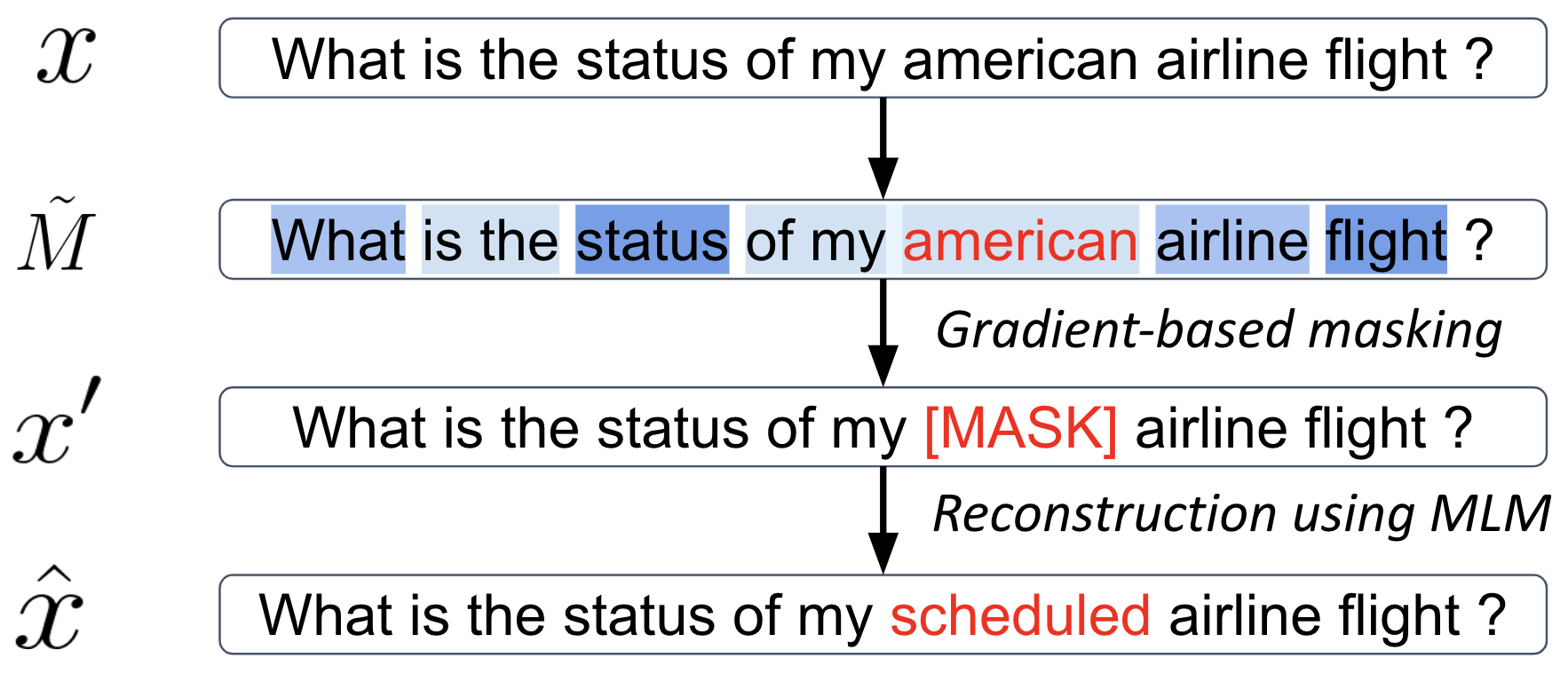}
		\caption{An illustrative example of GradAug. First, the smooth saliency $\tilde{M}$  is computed for each token, and we highlight important tokens in blue for the intent label ``flight\_status''. Less important tokens are more likely to be masked. Then, the masked token (``american'') is reconstructed by the MLM of BERT and the replacement token ``scheduled'' does not change the semantics of the original sentence. } \vspace{-0.05in}
		\label{fig:gradaug}
	\end{figure}

To utilize a pre-trained MLM (e.g. BERT) for text augmentation, the first step is to decide \emph{which tokens to mask}. Random sampling is used by the original BERT framework and a recent text augmentation method (SSMBA, \citet{ng2020ssmba}). However, if some crucial tokens are masked, the semantics might change after the reconstruction. For example, if the important token ``status'' in Figure \ref{fig:gradaug} is masked, top predictions from the MLM of BERT includes ``purpose'', ``cost'', and ``route'', which will potentially change the original semantics.

\paragraph{Gradient-based token masking.} Instead of randomly masking tokens, we compute a \emph{masking probability} $\bm{p} = [p_1, ..., p_n]$ for an input $x$ of $n$ tokens. 
For input $x$ with token embedding matrix\footnote{We use the token embeddings of BERT-like architectures, rather than position or segmentation embeddings.} $X= [X_1, ..., X_n]^\intercal \in \mathbb{R}^{n \times d}$  and label $y$, the \emph{importance} of tokens in $x$ to the label $y$ is computed by a \emph{saliency map} \cite{simonyan2013deep} $\bm{m}$:
\begin{equation}
\left\{
\begin{aligned}
& \bm{m} = \big [M(X_1), \dots, M(X_n) \big ]^\intercal   \in \mathbb{R}^{n}, \\ 
& M(X_i)  = \mathds{1}^\intercal \Bigg (\frac{\partial F^T_y(X)} {\partial X_i} \Bigg ) \in \mathbb{R}^{1},
\end{aligned}
\right.
\label{eq:vanilla_grad}
\end{equation}
where $F^T_y(X)$ is the \emph{Teacher} model's prediction score for the label $y$. $M(X_i)$  measures the \emph{importance} of the $i$-th token by accumulating the gradients of all elements in its embedding $X_i \in \mathbb{R}^{d}$ by differentiating $F^T_y(X)$ w.r.t. $X_i$. The intuition is that tokens with large gradients are important to the label $y$. However, previous studies \cite{sundararajan2017axiomatic,smilkov2017smoothgrad} pointed out that raw gradients can be very noisy and may sharply fluctuate locally. To this end, we compute a \emph{smooth saliency} measure \cite{smilkov2017smoothgrad} $\tilde{M}(X_i)$ for the $i$-th token as:
\begin{equation}
\left\{
\begin{aligned}
& \tilde{M}(X_i) = \frac{1}{m} \sum_{j=1}^m M(\tilde{X}_i^j) \in \mathbb{R}^{1}, \\
& \tilde{X}_i^j = X_i + \bm{z}^j,  
\end{aligned}
\right.
\label{eq:smooth_grad}
\end{equation}
where $m$ Gaussian noises $\bm{z}^j \sim \mathcal{N}(\bm{0}, \bm{\Sigma})  \in \mathbb{R}^{d} $ with mean $\mathbf{0}$ and diagonal co-variance matrix $\mathbf{\Sigma}$ are added to $X_i$ to calculate $m$ regular saliency measures, which average to the \emph{smooth saliency} $\tilde{M}(X_i)$ for $X_i$. 
The probability $p_i$ of masking the $i$-th token is inversely correlated to $\tilde{M}(X_i)$ as: 
\begin{equation}
p_i \propto \frac{1}{\tilde{M}(X_i)^\beta},
\label{eq:maskp}
\end{equation}
where $\beta$ controls the flatness of the distribution $\bm p$, and $\bm p$ is normalized by its sum. 
As the probability $p_i$ to mask a token $x_i$ is inversely correlated to its importance $\tilde{M}(X_i)$ to a downstream task, more important tokens are less likely to be masked.
We sample 15\%~\footnote{This is the default ratio used by BERT and SSMBA.} tokens of $x$ based on $\bm p$ and replace them by [MASK] to corrupt $x$ to $x'$. As $F^T$ is updated in each ST iteration, $\bm p$ is dynamically calculated in each ST iteration.

\begin{algorithm}[!t]
\caption{GradAug}\label{algo:gradaug}
\begin{algorithmic}[1]
\Require{Labeled data: $L$, Teacher: $F^T$, Number of augmentations per input: $q$ }
\Ensure{Augmented labeled data $L_{Aug}$}
\State Initialize $L_{Aug} \gets L$
\For{$\{x,y \} \in L$}
\State Compute \emph{masking probability} $\bm p$ using $F^T$ 
\For{$j \in 1 \dots q$}
\State $x' \gets$ Mask tokens of $x$ based on $\bm p$
\State $\hat{x} \gets$ Predict masked tokens by MLM
\State $L_{Aug}.append(\{\hat{x}, y\})$
\EndFor
\EndFor
\end{algorithmic}
\end{algorithm}

\paragraph{Reconstruction using MLM.} To reconstruct the masked tokens in $x'$, we utilize a pre-trained MLM to predict the [MASK] tokens. 
For stochastic purposes suggested by \citet{fan2018hierarchical}, we reconstruct each [MASK] by sampling 1 token from 10 most likely tokens according to their predicted probabilities.  Afterwards, we get a paraphrased $\hat{x}$ of the original $x$ as an augmentation. As our gradient-based masking scheme avoids replacing tokens crucial to the meaning of $x$, the label of $\hat{x}$ is preserved the same as $x$. 

An illustrative example of GradAug is given in Figure \ref{fig:gradaug}, and the detailed procedure applying GradAug on $L$ is described in Algorithm \ref{algo:gradaug}.

%% file: text/exp.tex
\section{Experiments}

\input{tables/intent-main.tex}

\subsection{Dataset Description}
We evaluate four different datasets for four downstream tasks as in \citet{wu2020tod}.

\textbf{OOS} \cite{larson2019evaluation} is a benchmark dataset for intent classification in ToD. It consists of 150 in-domain intents and 1 out-of-scope intent. The full dataset contains 15,100/3,100/5,500 samples for train/validation/test, and all data are balanced across 151 different intents. 

\textbf{MWOZ} \cite{eric2020multiwoz} is evaluated in three downstream tasks, including dialog state tracking, dialogue act prediction, and response prediction. It contains 8,420/1,000/1,000 dialogues for train/validation/test. For dialog act prediction, we remove the domain information from original labels as in \citet{wu2020tod}, resulting 13 DA intents.

\textbf{DSTC2} \cite{henderson2014second} and \textbf{GSIM} \cite{shah2018bootstrapping} are two corpus used in dialog act prediction and response selection tasks. DSTC2 contains 1,612/506/1,117 dialogues for train/validation/test; GSIM contains 1,500/469/1,039 dialogues for train/validation/test. DA intent labels of DSTC2 and GSIM are mapped to universal dialogue acts \cite{paul2019towards}, resulting in 19 and 13 DA intents respectively. 

\subsection{Experiment Settings}

We randomly sample 1\% or 10\% of the training data to serve as the initial labeled data $L$, while the remainders are used as unlabeled data $U$.
We report mean and standard deviation with three different random seeds for each experiment to reduce data sampling variance. We also report the upper bound of pre-trained models \emph{without} ST using all labeled training data, referred to as ``Full''.

We test two pre-trained models: (i). uncased base BERT with 110M parameters; (ii). ToD-BERT~\footnote{We used their joint version (ToD-BERT-jnt) pre-trained with the MLM and ``response contrastive loss'' objectives.} \cite{wu2020tod} that is further pre-trained on 9 public ToD datasets on top of BERT. When ST is applied to them, the corresponding MLM is used by GradAug to reconstruct masked tokens. Basic model parameters of the first 3 downstream tasks are set the same as \citet{wu2020tod}. In response selection, we reduced training batch size from 25 to 20 to fit our computation constraint.

BERT and Tod-BERT without ST are trained on the initial labeled data $L$ until validation performance does not improve for 20 epochs~\footnote{Our different (often better) results compared to the ToD-BERT paper mainly come from this stricter early stop criteria.}.
For ST, when the \emph{Student} is trained on $L_{Aug}$ in one ST iteration (c.f. Algorithm \ref{algo:ST} line 12), we apply early stop until validation performance does not improve for 10 epochs. Moreover, the best \emph{Student} across multiple ST iterations is selected based on validation performance (c.f. Algorithm \ref{algo:ST} line 2). It means that the best \emph{Student} model does \textit{not} necessarily use up all unlabeled data. Other hyper-parameters of ST selected base on validation performance are reported in Appendix \ref{appendix:parameter}.


\subsection{Main Results of Four Downstream Tasks}

\input{tables/dst-main.tex}

\input{tables/da-main.tex}
\input{tables/rs-main.tex}

\quad \textbf{Intent classification}. Results of intent classification on OOS are presented in Table~\ref{tab:intent-main} with \emph{accuracy} of \textbf{all} 151 intents; 150 \textbf{in}-domain intents; the \textbf{out}-of-scope intent, and the \emph{recall} of the out-of-scope intent. 
ST significantly improves the pre-trained BERT and ToD-BERT. 
When only 1\% labeled data are used, ST achieves 33.6\% and 36.8\% higher accuracy on all 151 intents for BERT and ToD-BERT respectively. For 10\% labeled data, the above two margins are 7.0\% and 9.8\%.
Furthermore, ST largely improves the recall of the out-of-scope intent, indicating that it is more robust to out-of-scope intents with noisy distributions.

\textbf{Dialog state tracking}. Results of dialog state tracking on MWOZ are presented in Table~\ref{tab:dst-main}. Two common evaluation metrics \cite{budzianowski2018multiwoz,wu2019transferable} are used: \emph{slot accuracy} and \emph{joint goal accuracy}. Slot accuracy is computed for each individual state (domain, slot, value) to check whether the value is correctly predicted. Joint goal accuracy checks whether the predicted states exactly matches the
ground truth states. We could see that ST consistently improves both BERT and ToD-BERT. E.g., ST has 1.5\% and 2.8\% joint goal accuracy improvement over ToD-BERT when 1\% and 10\% labeled data are used respectively. Similar margins can be observed for ST on top of BERT.

\textbf{Dialog act prediction}. Experiments are conducted on three datasets and results are reported in Table~\ref{tab:da-main}. We report \emph{micro-F1} and \emph{macro-F1} scores for this multi-label classification task. Again, the benefit of ST can be observed by the improvement for both BERT and ToD-BERT. When 10\% labeled data are used, BERT and ToD-BERT perform similarly to their upper bound (Full), and the improvement margin of ST is limited. When 1\% labeled data are used, more notable margins of ST can be seen on the two simpler datasets (DSTC2, GSIM) and the macro-F1 score of MWOZ.

\textbf{Response selection}. Results of response selection on three datasets are reported in Table~\ref{tab:rs-main}. We randomly sample 100 responses as negative responses and report Recall@1\&3 \cite{henderson2019training} indicating whether the true response is ranked in the top-1 or top-3 predicted responses. 
When 1\% labeled data are used, ST achieves 6\%, 12.3\%, and 16.4\% higher Recall@1 accuracy over ToD-BERT on three datasets respectively. For 10\% labeled data, the three margins above are 13.0\%, 7.5\%, and 14.4\% respectively. Larger improvements can be observed for ST on top of BERT.

Altogether, our experiments on four different downstream tasks reveal that: 
\begin{itemize}[itemsep=-1pt,topsep=2pt,leftmargin=12pt]
\item Self-training provides complementary benefits on top of pre-training. ST consistently improves both BERT and ToD-BERT on all four downstream tasks with only 1\% and 10\% labeled data. 
\item Self-training is on par with customized pre-training for ToD. BERT performs worse than ToD-BERT, yet BERT-ST achieves comparable or even better performance than ToD-BERT which is heavily pre-trained on ToD corpora.
\item Self-training bridges the gap between few-shot learning and full supervision. BERT and ToD-BERT with 10\% labeled data perform much worse than models using all labeled data (“Full”) for intent classification and response selection. ST largely improves performances in these two cases with results comparable to ``Full''.
\item The benefit of self-training is evident on two simpler single-label prediction tasks (intent classification, response selection), indicated by 6-37\% gain with 1\% labeled data; 7-15\% gain with 10\% labeled data.  The margin is smaller on two other more challenging multi-label prediction tasks. 
\end{itemize}

\subsection{In-depth Analyses of Self-training}

In this section, we provide in-depth analyses of the proposed self-training approach. As case studies, we limited our discussion on intent classification (\textbf{IC}) on OOS and response selection (\textbf{RS}) on GSIM using ToD-BERT-ST with 10\% labeled data. Reported results are accuracies on all intents and Recall@3 respectively.

\input{tables/ablation.tex}

\paragraph{Ablation study.} In Table~\ref{tab:ablation}, we compare three simplified versions of ToD-BERT-ST to understand the effects of different components. We can observe that: (i) Masking tokens using the smooth saliency computed in Eq. (\ref{eq:smooth_grad}) for GradAug is beneficial because replacing it by the vanilla saliency in Eq. (\ref{eq:vanilla_grad}) (“w/o Smooth Saliency”) degrades the performance by 3.4\% and 0.5\% on IC and RS. (ii) Training a more robust \emph{Student} using data augmented by GradAug is advantageous because dropping this augmentation step (“w/o Augmentation”) impairs performance by 4.9\% and 10.1\%. (iii) The Pseudo-Labeling operation to iteratively label unlabeled data is important for ST, indicated by the 8.4\% and 15.2\% performance drop of ``w/o Pseudo-Labeling'' that only applies GradAug to the initial labeled data without utilizing unlabeled data.

\paragraph{Comparison to other \emph{Selectors} in ST.} In Table \ref{tab:st-compare}, we compare our scheme of selecting samples with top-$k$ confident predictions from $U$ in each iteration with (i) Random-$k$: randomly select $k$ samples; (ii) Least-$k$: select samples with least-$k$ confident predictions (iii) Select-all \cite{xie2020self,du2020self}: label all samples of $U$ in an iteration and relabel them in the next iteration. We could see that ``Random-$k$'' and ``Least-$k$'' perform worse than ours, yet they both outperform ``Select-all'' by large margins. It means that the initial \emph{Teacher} trained on limited labeled data is not good enough to assign reliable labels to a large number of unlabeled data.

\input{tables/st-compare.tex}
\input{tables/aug-comp.tex}

\paragraph{Comparison to other text augmentation methods.}
In Table~\ref{tab:aug-comp}, we compare GradAug with four representative text augmentation methods to augment $L$. 
We follow the default setting of these techniques and apply them to our ST pipeline to generate three paraphrases for each input as in GradAug. We could see that GradAug consistently outperforms the current state-of-the-art (SSMBA, MixText, UDA), and it outperforms EDA by large margins. As EDA might easily change the input semantics, it even performs worse than using no data augmentation for intent classification. This result reinforces the importance of preserving semantics during augmentation for ToD.

%% file: tables/intent-main.tex
\begin{table*}[t!]
\centering
\small
\begin{tabular}{l|l|c|c|c|c}
\hline
\multirow{2}{*}{\bf Data} & \multirow{2}{*}{\bf Model}    & \bf Acc. & \bf Acc. & \bf Acc. & \bf Recall \\
 &    & \bf (all) & \bf (in) & \bf (out) & \bf (out) \\
\hline
\multirow{4}{*}{1\%}     &  BERT        & 36.5\%  $\pm$ 2.7\% &  44.6\% $\pm$ 3.3\% & 81.8\% $\pm$ 0.8\%  &  0.4\%  $\pm$ 0.2\%\\
                         &  BERT-ST     &  \textbf{70.1\%}  $\pm$ 2.2\%  &   \textbf{82.2\%}  $\pm$ 3.7\%   &  \textbf{84.3\%}  $\pm$ 1.0\%   &  \textbf{15.5\%} $\pm$ 1.3\% \\
                         &  ToD-BERT    & 39.0\%  $\pm$ 1.3\% &  47.1\%  $\pm$ 0.7\% &  82.0\%  $\pm$ 0.4\% &  2.3\% $\pm$ 0.3\% \\
                         &  ToD-BERT-ST & \textbf{75.8\%} $\pm$ 1.7\%  & \textbf{87.8\%}  $\pm$ 1.5\% &  \textbf{85.5\%}  $\pm$ 0.7\% & \textbf{21.9\%} $\pm$ 0.9\%  \\
\hline
\multirow{4}{*}{10\%}    &  BERT        &  73.6\% $\pm$ 1.9\%   &  87.4\% $\pm$ 2.1\% &  83.9\% $\pm$ 0.5\% & 11.7\% $\pm$ 1.1\% \\
                         &  BERT-ST     &  \textbf{80.6\%}  $\pm$ 1.7\%  & \textbf{94.3\%} $\pm$ 1.5\%  & \textbf{84.9\%}  $\pm$ 0.6\%    &  \textbf{17.1\%}  $\pm$ 0.9\% \\
                         &  ToD-BERT    & 75.5\%  $\pm$ 1.0\% & 89.4\% $\pm$ 0.8\% & 84.1\% $\pm$ 0.7\% &  13.3\%  $\pm$ 1.4\%\\
                         &  ToD-BERT-ST &  \textbf{85.3\%} $\pm$ 0.9\%   &  \textbf{94.7\%} $\pm$ 0.7\%   &  \textbf{89.4\%} $\pm$ 0.6\% & \textbf{42.8\%} $\pm$ 1.7\%  \\
\hline
\multirow{2}{*}{Full*} &  BERT    &  84.9\% &  95.8\%   &  88.1\%   &    35.6\%  \\
 &  ToD-BERT    &  86.6\% &  96.2\%   &  89.9\%   &    43.6\%  \\
\hline
\end{tabular}%
\caption{\label{tab:intent-main} Results of intent classification. Bold numbers indicate ST improves the corresponding pre-trained model. Results with * are taken from \citet{wu2020tod}.} \vspace{-0.1in}
\end{table*}

%% file: tables/dst-main.tex
\begin{table}[!]
\centering
\small
\setlength{\tabcolsep}{4pt}
\begin{tabular}{l|l|c|c}
\hline
\bf Data & \bf Model  & \bf Joint Acc & \bf Slot Acc \\
\hline
\multirow{4}{*}{1\%}  &  BERT & 8.0\%  $\pm$ 1.1\% &  84.3\% $\pm$ 0.6\% \\
&  BERT-ST     &  \textbf{8.8\%} $\pm$ 0.6\% &  \textbf{84.5\%} $\pm$ 0.4\% \\
&  ToD-BERT     & 8.4\%  $\pm$ 0.5\% & 85.7\% $\pm$ 0.4\%\\
&  ToD-BERT-ST & \textbf{9.9\%}  $\pm$ 0.3\% &  \textbf{86.5\%}  $\pm$ 0.2\%   \\
\hline
\multirow{4}{*}{10\%}  &  BERT  & 21.2\%  $\pm$ 0.5\% & 92.0\%  $\pm$ 0.3\%  \\
&  BERT-ST     & \textbf{23.9\%} $\pm$ 0.3\% & \textbf{92.4\%} $\pm$ 0.5\% \\
&  ToD-BERT    & 25.5\%  $\pm$ 0.6\%  &  93.4\%  $\pm$ 0.2\%  \\
&  ToD-BERT-ST &  \textbf{28.3\%} $\pm$ 0.4\% & \textbf{93.7\%}  $\pm$ 0.1\% \\
\hline
\multirow{2}{*}{Full*} &  BERT    & 45.6\%     &    96.6\%      \\
 &  ToD-BERT    & 48.0\%     &    96.9\%      \\
\hline
\end{tabular}%
\caption{\label{tab:dst-main} Results of dialog state tracking. Bold numbers indicate ST improves the corresponding pre-trained model. Results with * are taken from \citet{wu2020tod}.} \vspace{-0.1in}
\end{table}

%% file: tables/da-main.tex
\begin{table*}[htb!]
\centering
\small
\setlength{\tabcolsep}{2.9pt}
\begin{tabular}{c|l|c|c|c|c|c|c}
\hline
\multicolumn{1}{l|}{\multirow{2}{*}{\bf Data}} & \multirow{2}{*}{\bf Model} & \multicolumn{2}{c|}{\bf MWOZ}  & \multicolumn{2}{c|}{\bf DSTC2}                                    & \multicolumn{2}{c}{\bf GSIM}    \\ 
\cline{3-8} 
\multicolumn{1}{l|}{}   &   & \multicolumn{1}{c|}{micro-F1} & \multicolumn{1}{c|}{macro-F1} & \multicolumn{1}{c|}{micro-F1} & \multicolumn{1}{c|}{macro-F1} & \multicolumn{1}{c|}{micro-F1} & \multicolumn{1}{c}{macro-F1} 
\\ 
\hline
\multirow{4}{*}{1\%} & BERT & 83.5\% $\pm$ 0.7\% & 61.2\% $\pm$ 1.5\% & 79.1\%  $\pm$ 1.4\% & 26.8\% $\pm$ 0.4\% &  70.3\%   $\pm$ 1.3\%& 27.9\%    $\pm$ 0.7\%          \\
 & BERT-ST   & 82.7\% $\pm$ 0.5\% & \textbf{64.2\%} $\pm$ 0.8\% & \textbf{81.4\%} $\pm$ 0.7\% & \textbf{27.3\%} $\pm$ 0.2\% & \textbf{73.0\%}   $\pm$ 0.8\% &\textbf{29.8\%} $\pm$ 0.4\% \\
 & ToD-BERT  & 85.8\%  $\pm$ 0.2\% & 67.0\% $\pm$ 0.6\% & 80.9\%  $\pm$ 0.7\% & 25.3\% $\pm$ 0.4\% & 86.5\%  $\pm$ 3.0\% & 36.6\% $\pm$ 2.1\% \\
 & ToD-BERT-ST   & \textbf{86.9\%} $\pm$ 0.4\% & \textbf{71.8\%}   $\pm$ 0.3\% & \textbf{82.7\%} $\pm$ 0.8\%  & \textbf{28.5\%} $\pm$ 0.4\% & \textbf{92.6\%} $\pm$ 1.1\% & \textbf{40.8\%}     $\pm$ 0.9\%                   \\ \hline
\multirow{4}{*}{10\%}   & BERT   & 89.8\%   $\pm$ 0.2\% & 77.8\% $\pm$ 0.3\% & 88.9\% $\pm$ 0.7\% & 35.7\%  $\pm$ 1.3\% & 97.1\%  $\pm$ 0.3\% & 44.1\%  $\pm$ 0.2\% \\
& BERT-ST  & 89.5\% $\pm$ 0.1\% & \textbf{79.2\%} $\pm$ 0.6\% & \textbf{92.3\%} $\pm$ 0.6\% & \textbf{38.4\%}  $\pm$ 1.0\% &  \textbf{97.6\%} $\pm$ 0.2\% & \textbf{44.6\%} $\pm$ 0.4\% \\
& ToD-BERT & 90.0\% $\pm$ 0.2\% & 78.4\% $\pm$ 1.0\% & 90.6\% $\pm$ 2.1\% & 38.8\% $\pm$ 1.9\%  & 98.6\%  $\pm$ 0.2\% & 44.9\% $\pm$ 0.2\%\\
& ToD-BERT-ST & \textbf{90.2\%} $\pm$ 0.2\% & \textbf{79.6\%}   $\pm$ 0.4\%& \textbf{92.9\%} $\pm$ 0.8\%  & \textbf{40.5\%} $\pm$ 0.9\% & \textbf{99.3\%} $\pm$ 0.3\% & \textbf{45.6\%}  $\pm$ 0.4\%  \\ \hline
\multirow{2}{*}{Full*}    & BERT & 91.4\% & 79.7\%  & 92.3\%  & 40.1\% & 98.7\%   & 45.2\%   \\ 
   & ToD-BERT & 91.7\% & 80.6\%  & 93.8\%  & 41.3\% & 99.5\%  & 45.8\%   \\ 
\hline
\end{tabular}%
\caption{\label{tab:da-main} Results of dialog act prediction. Bold numbers indicate ST improves the corresponding pre-trained model. Results with * are taken from \citet{wu2020tod}.}
\end{table*}

%% file: tables/rs-main.tex
\begin{table*}[htb!]
\centering
\small
\setlength{\tabcolsep}{2.8pt}
\begin{tabular}{c|l|c|c|c|c|c|c}
\hline
\multicolumn{1}{l|}{\multirow{2}{*}{  \textbf{Data}}} & \multirow{2}{*}{  \textbf{Model}} & \multicolumn{2}{c|}{  \textbf{MWOZ}}    & \multicolumn{2}{c|}{  \textbf{DSTC2}}  & \multicolumn{2}{c}{  \textbf{GSIM}}   \\ 
\cline{3-8} 
\multicolumn{1}{l|}{}   &    & \multicolumn{1}{c|}{Recall@1} & \multicolumn{1}{c|}{Recall@3} & \multicolumn{1}{c|}{Recall@1} & \multicolumn{1}{c|}{Recall@3} & \multicolumn{1}{c|}{Recall@1} & \multicolumn{1}{c}{Recall@3} 
\\ 
\hline
\multirow{4}{*}{1\%}    & BERT   & 7.3\%  $\pm$ 1.4\%  & 19.5\% $\pm$ 3.2\% & 3.5\% $\pm$ 0.6\%  & 9.8\% $\pm$ 1.5\%     & 4.0\% $\pm$ 0.6\%  & 11.4\% $\pm$ 1.0\%   \\
& BERT-ST    & \textbf{23.8\%} $\pm$ 1.4\% & \textbf{46.1\%} $\pm$ 0.7\% & \textbf{36.7\%} $\pm$ 0.4\% &  \textbf{51.1\%} $\pm$ 1.3\% &   \textbf{11.1\%} $\pm$ 0.8\% &   \textbf{24.2\%} $\pm$ 0.6\% \\
& ToD-BERT  & 37.5\% $\pm$ 1.9\% & 63.0\% $\pm$ 1.1\% & 35.7\%  $\pm$ 0.9\%  & 53.8\% $\pm$ 0.7\%  & 11.4\%  $\pm$ 1.1\% & 24.1\%   $\pm$ 0.9\%   \\ 
& ToD-BERT-ST  &   \textbf{43.5\%} $\pm$ 0.7\% &   \textbf{66.3\%} $\pm$ 0.6\% &   \textbf{48.0\%} $\pm$ 0.5\% &   \textbf{64.6\%} $\pm$ 0.3\% &  \textbf{27.8\%} $\pm$ 1.0\%& \textbf{42.9\%} $\pm$ 0.8\%\\ \hline
\multirow{4}{*}{10\%} & BERT & 26.1\% $\pm$ 3.0\% & 56.5\% $\pm$ 3.5\% & 27.7\% $\pm$ 2.1\% & 42.9\% $\pm$ 3.4\% & 13.4\% $\pm$ 0.6\% & 28.3\% $\pm$ 1.7\% \\
& BERT-ST  &   \textbf{43.1\%} $\pm$ 1.1\% &  \textbf{66.1\%} $\pm$ 1.3\% &   \textbf{53.7\%}$\pm$  2.0\%&   \textbf{67.1\%} $\pm$ 2.9\%&  \textbf{22.3\%} $\pm$ 0.4\% &  \textbf{40.4\%} $\pm$ 0.9\%\\
& ToD-BERT    & 47.2\% $\pm$ 1.1\% & 69.4\% $\pm$ 1.1\% & 51.3\% $\pm$ 0.7\% & 66.0\%  $\pm$ 0.49\% & 28.5\% $\pm$ 0.7\% & 47.8\% $\pm$ 1.0\% \\
& ToD-BERT-ST &  \textbf{60.2\%} $\pm$ 1.3\% &   \textbf{81.9\%}  $\pm$ 1.6\%  &   \textbf{58.8\%}  $\pm$ 0.8\%   &   \textbf{72.2\%}  $\pm$ 1.1\%       &  \textbf{41.8\%} $\pm$ 0.9\%  &   \textbf{64.9\%} $\pm$ 1.4\%  \\ \hline
\multirow{2}{*}{Full} & BERT & 47.5\% & 75.5\% & 46.6\% & 62.1\% & 13.4\%  & 32.9\%   \\
  & ToD-BERT & 66.9\% & 89.1\% & 59.5\% & 73.1\% & 43.0\%  & 65.3\%   \\
\hline
\end{tabular}%

\caption{\label{tab:rs-main} Results of response selection. Bold numbers indicate ST improves the corresponding pre-trained model.}
\vspace{-0.08in}
\end{table*}

%% file: tables/ablation.tex


\begin{table}[t!]
\centering
\begin{tabular}{l|c|c}
\hline
& \bf IC  & \bf  RS \\
& \bf Acc. (all)  & \bf Recall@3\\
\hline
ToD-BERT-ST & 85.3\% & 64.9\% \\
w/o Smooth Saliency & 81.9\% & 64.4\% \\
w/o Augmentation & 80.4\% & 54.8\% \\
w/o Pseudo-Labeling & 76.9\% & 49.7\%\\
\hline
ToD-BERT &  75.5\% &  47.8\%\\
\hline
\end{tabular}
\caption{\label{tab:ablation}Ablation study of ST for intent classification (IC) on OOS and response selection (RS) on GSIM.} \vspace{-0.1in}
\end{table}

%% file: tables/st-compare.tex



\begin{table}[t!]
\centering

\begin{tabular}{l|c|c}
\hline
& \bf IC  & \bf  RS\\
& \bf Acc. (all)  & \bf Recall@3\\
\hline
Top-$k$ (Ours) & 85.3\% & 64.9\%\\
Random-$k$ & 84.0\% & 64.1\%\\
Least-$k$ & 82.7\%  & 61.4\%\\
Select-all & 76.0\%  & 50.8\%\\
\hline
\end{tabular}

\caption{\label{tab:st-compare}Comparison to other \emph{Selectors} in ST for intent classification (IC) on OOS and response selection (RS) on GSIM.}
\end{table}

%% file: tables/aug-comp.tex



\begin{table}[t!]
\centering
\setlength{\tabcolsep}{1.5pt}
\begin{tabular}{l|c | c}
\hline
& \bf IC  & \bf  RS \\
& \bf Acc. (all)  & \bf Recall@3\\
\hline
GradAug (Ours) & 85.3\% & 64.9\%\\
SSMBA \cite{ng2020ssmba} & 84.6\% & 64.2\% \\
MixText \cite{chen2020mixtext} & 83.6\% & 62.7\%\\
UDA \cite{xie2019unsupervised} &  82.5\% & 62.2\%\\
EDA \cite{wei2019eda} &  77.2\% &  57.6\% \\
\hline
w/o Augmentation & 80.4\% &  54.8\% \\
\hline
\end{tabular}

\caption{\label{tab:aug-comp}Comparison to other text augmentation methods to train the \emph{Student} for intent classification (IC) on OOS and response selection (RS) on GSIM.} 
\end{table}

%% file: text/conclusion.tex
\section{Conclusion}

We study using self-training to improve the strong pre-trained models for few-shot learning tasks in ToD. 
An iterative self-training method with a new text augmentation technique (GradAug) is proposed to gradually train a stronger \emph{Student} model using unlabeled data.
Extensive empirical results on four downstream tasks in ToD demonstrate the consistent improvements of self-training on top of pre-trained models. 
Our findings on using self-training to improve learning from limited labeled data may inspire future studies towards building more sample-efficient and scalable ToD systems.

%% file: tables/dataset-setting.tex
\begin{table*}[h]
\setlength{\tabcolsep}{4pt}
\centering
\begin{tabular}{c|l|c|c|c}
\hline
\bf Task                 & \bf Dataset                & \bf Ratio of Labeled Data & \bf Initial Samples in \emph{L} & \bf Hyper-parameter \emph{k} \\ \hline
\multirow{2}{*}{Intent classification}  & \multirow{2}{*}{OOS}   & 1\% & 151  & 500 \\
                     &                        & 10\%      & 1510    &  1510   \\ \hline
\multirow{2}{*}{Dialog state tracking} & \multirow{2}{*}{MWOZ}  & 1\% & 567  & 200 \\
                     &                        & 10\%      &   5672   &  2000      \\ \hline
\multirow{6}{*}{Dialog act prediction}  & \multirow{2}{*}{MWOZ} & 1\% &  482  & 500\\
                     &                        & 10\%  & 4824   &  2000    \\ \cline{2-5} 
                     & \multirow{2}{*}{DSTC2} & 1\%     & 116  &  200   \\
                     &                        & 10\%  & 1160   &  500       \\ \cline{2-5} 
                     & \multirow{2}{*}{GSIM}  & 1\%    & 66    &  100   \\
                     &                        & 10\%   & 664   &  500        \\ \hline
\multirow{6}{*}{Response selection} & \multirow{2}{*}{MWOZ} & 1\% & 482 & 500 \\
                     &                        & 10\%   & 4824  & 4500   \\ \cline{2-5} 
                     & \multirow{2}{*}{DSTC2} & 1\%   & 100    &  100  \\
                     &                        & 10\%  & 1006   &  700  \\ \cline{2-5} 
                     & \multirow{2}{*}{GSIM}  & 1\%   &  66    &   80   \\
                     &                        & 10\%   &  664   &  500    \\ \hline
\end{tabular}
\caption{\label{tab:dataset-setting}Dataset specifications for experiments in different downstream tasks and the hyper-parameter \emph{k} (\textbf{the rightmost column}) indicating the number of pseudo-labeled data in each self-training iteration in different experiments for four downstream tasks.}
\end{table*}

%% file: text/appendix.tex
\appendix

\part*{ Appendix}

\section{Reproducibility Checklist}

\subsection{Code}
Our code will be available at \url{https://github.com/MiFei/ST-ToD} soon.

\subsection{Dataset Specifications for Different Tasks}

The exact dataset scales regarding the initial 1\% or 10\% labeled data in different few-shot learning scenarios for different downstream tasks are reported in Table~\ref{tab:dataset-setting} in the column headed with ``\textbf{Initial Samples in $L$}''.

\subsection{Hyper-parameters of ST}
\label{appendix:parameter}
\input{tables/hyperparams}

The number ($k$) of pseudo-labeled data in each ST iteration in each experiment setting is reported in the rightmost column of Table~\ref{tab:dataset-setting}.
Other hyper-parameters of ST are reported in Table~\ref{tab:hyperparam}, and they are shared across different downstream tasks and datasets.

An exhaustive search on ST hyper-parameters is not conducted because it is very expensive to finely tune large pre-trained models on all four downstream tasks for different datasets. Therefore, we fix $q, m$ and manually tune other hyper-parameters within reasonable ranges around current values indicated in Table~\ref{tab:dataset-setting} and Table~\ref{tab:hyperparam}.
We could expect that even better results of ST can be achieved with a thorough hyper-parameter search by researchers or practitioners without computation constraints.
To provide more insight, we found that setting $k$ too small compromises computation time, while setting it too large compromises performance.

All experiments are conducted using a single GPU (GTX TITAN X), and eight GPUs are used in total.

%% file: tables/hyperparams.tex
\begin{table}[!htb]
\centering
\setlength{\tabcolsep}{2.5pt}
\begin{tabular}{l|l|l|l}
\hline
  & \bf Source & \bf Definition & \bf Value \\ \hline
$q$ &  Alg.~\ref{algo:gradaug}  & \# of augmentations per input & 3 \\
$\beta$  &  Eq.~\ref{eq:maskp} &  flatness of distribution $\bm p$ & 1.0 \\
$m$ &  Eq.~\ref{eq:smooth_grad} & \# of Gaussian noises & 20 \\
$\bm \Sigma$  &  Eq.~\ref{eq:smooth_grad} & diagonal co-variance matrix & 1e-4$ \cdot \bm I$  \\
\hline
\end{tabular}
\caption{\label{tab:hyperparam}Hyper-parameters of ST that are shared across different downstream tasks and datasets.}
\end{table}
